\documentclass[preprint, prX]{revtex4}

\usepackage{amsmath}    % need for subequations
\usepackage{graphicx}   % need for figures
\usepackage{verbatim}   % useful for program listings
\usepackage{color}      % use if color is used in text
\usepackage{subfigure}  % use for side-by-side figures
\usepackage{hyperref}   % use for hypertext links, including those to external documents and URLs
\usepackage{amssymb}    % Necessary for \Bbb{R}, etc.
\usepackage{epsfig}
\usepackage{graphics,graphicx}
\usepackage{setspace}
\usepackage{url}
\usepackage{algorithm,algorithmic}
\usepackage{multirow}

\usepackage{amsfonts}
\usepackage{amssymb}
\usepackage{latexsym}
\usepackage{sidecap}

\begin{document}

\begin{abstract}

Differential evolution possesses a multitude of various strategies for generating new trial solutions. Unfortunately, the best strategy is not known in advance. Moreover, this strategy usually depends on the problem to be solved. This paper suggests using various regression methods (like random forest, extremely randomized trees, gradient boosting, decision trees, and a generalized linear model) on ensemble strategies in differential evolution algorithm by predicting the best differential evolution strategy during the run. Comparing the preliminary results of this algorithm by optimizing a suite of five well-known functions from literature, it was shown that using the random forest regression method substantially outperformed the results of the other regression methods.

Citation details:  Fister, I. Jr., Fister, I., Brest, J.  Comparing various regression methods on ensemble strategies in differential evolution, In Proceedings of 19th International Conference on Soft Computing MENDEL 2013, Brno, 2013, pp. 87-92.

\end{abstract}

\title{Comparing various regression methods on ensemble strategies in differential evolution}

\author{Iztok Fister Jr.}
\altaffiliation{University of Maribor, Faculty of electrical engineering and computer science
Smetanova 17, 2000 Maribor}
\email{iztok.fister@guest.arnes.si}

\author{Iztok Fister}
\altaffiliation{University of Maribor, Faculty of electrical engineering and computer science
Smetanova 17, 2000 Maribor}
\email{iztok.fister@uni-mb.si}

\author{Janez Brest}
\altaffiliation{University of Maribor, Faculty of electrical engineering and computer science
Smetanova 17, 2000 Maribor}
\email{janez.brest@uni-mb.si}

\maketitle

\section{Introduction}

Nowadays, differential evolution (DE) has probably become one the most popular and effective evolutionary algorithms used for solving optimization problems. This method was introduced in 1995 by Storn and Price~\cite{storn1997differential}. It is a population-based algorithm, where each individual (i.e., real-valued vector) represents a candidate solution for the problem to be solved. This solution undergoes the effects of mutation and crossover operators, and thereby generate a population of trial solutions. Each trial solution competes with its corresponding candidate solutions for a position in a new population. The selection operator is responsible for selecting the best between the trial and candidate solutions by judging the quality of solution according to the value of the fitness function.

Three algorithms' parameters control the behavior of DE: the amplification factor $F$, the crossover parameter $\mathit{CR}$, and the population size $\mathit{NP}$. In the original DE, these parameters are fixed during the optimization search process. However, these parameters that may be good at the beginning of the search process but may become worse in later generations or vice versa. Therefore, several adaptive and self-adaptive algorithms have been arisen that are able to modify the algorithms' parameters during the run, for example, jDE~\cite{brest2006self}, SaDE~\cite{qin2005self}. On the other hand, methods using the combination of various ensemble of parameters and mutation strategies~\cite{epsde,Tvrdik:2006,Tvrdik:2007}, have been improved the behavior of the classical DE algorithm significantly. A complete review of DE methods can be found in~\cite{das2011differential,Tvrdik:2009}. 

Making decisions based on the input of multiple experts has been a common practice in human civilization~\cite{Polikar:2012}. The computational intelligence (CI) and machine learning (ML) community have studied methods that share such a joint-decision procedure. In line with this, ensemble learning has emerged that introduces robustness and accuracy in a decision process. Ensemble learning can be applied to many real-world applications. In our study, ensemble learning was used for predicting the more appropriate regression real-valued vector obtained from an ensemble of multiple DE strategies. In place of the ordinal offspring, the predicted regression vector was entered into the original DE selection process. The following ensemble learning methods are taken into account: random forest (RF), extremely randomized trees (EXT), and gradient boosting (GB). In order to complete our comparisons, some older regression methods were also included into this study, like decision trees (DT), and a generalized linear model (LM). Note that all these methods were implemented in Python using the scikit-learn python library~\cite{scikit-learn}.

The proposed algorithm was tested by optimizing a suite of five well-known functions taken from literature. The preliminary results using mentioned regression methods on ensemble strategies in DE algorithm showed that the best predictions were performed by the RF regression method.

The structure of this paper is as follows: In Section 2, the background information is discussed as needed for a better understanding the proposed DE algorithm. The proposed algorithm is presented in Section 3. The experiments and results are illustrated in Section 4. In Section 5, the conclusions are obtained and directions for further work are sketched.

\section{Background}
\label{DE algorithm}

This section deals with the background information needed for understanding the proposed DE algorithm. Firstly, an original DE algorithm is described. In line with this, each strategy from the ensemble strategies is enumerated and discussed in detail. Then, the regression methods used for predicting the best strategy during the run are briefly mentioned (e.g., random forest, extremely randomized trees, gradient boosting, decision trees, and a generalized linear model). Note that these methods are implemented in standard libraries (e.g., in~\cite{scikit-learn}) and therefore, no special efforts were needed for implementation. 

\subsection{Differential evolution}

Differential evolution (DE)~\cite{das2011differential} is an evolutionary algorithm appropriate for continuous and combinatorial optimization that was introduced by Storn and Price in 1995~\cite{storn1997differential}. This is a population-based algorithm that consists of $\mathit{NP}$ real-valued vectors representing the candidate solutions, as follows:

\begin{equation}
%\vspace{-1mm}
\label{eq:de_repr}
 \mathbf{x}_{i}^{(t)}=(x_{i1}^{(t)}, \ldots ,x_{iD}^{(t)}),\ \ \ \textnormal{for}\ i=1 \ldots \mathit{NP},
%\vspace{-2mm}
\end{equation}

\noindent where $D$ denotes the dimension of the problem. 

The DE supports a differential mutation, a differential crossover and a differential selection. In particular, the differential mutation randomly selects two solutions and adds a scaled difference between these to the third solution. This mutation can be expressed as follows:

\begin{equation}
%\vspace{-1mm}
\label{eq:de_mut}
 \mathbf{u}_{i}^{(t)}=\mathbf{x}_{r0}^{(t)}+F \cdotp (\mathbf{x}_{r1}^{(t)}-\mathbf{x}_{r2}^{(t)}),\ \ \ \textnormal{for}\ i=1 \ldots \mathit{NP},
%\vspace{-2mm}
\end{equation}

\noindent where $F \in [0.1,1.0]$ (Price and Storn proposed $F \in [0.0,2.0]$, but normally this interval is used in the DE community) denotes the scaling factor as a positive real number that scales the rate of modification whilst $r0,\ r1,\ r2$ are randomly selected values in the interval $1 \ldots \mathit{NP}$. 

Uniform crossover is employed as a differential crossover by the DE. The trial vector is built from parameter values copied from two different solutions. Mathematically, this crossover can be expressed as follows:

\begin{equation}
%\vspace{-1mm}
\label{eq:de_xover}
 w_{i,j}=\begin{cases}
          u_{i,j}^{(t)} & \textnormal{rand}_{j}(0,1) \leq \mathit{CR} \vee j=j_{rand}, \\
		  x_{i,j}^{(t)} & \text{otherwise} ,
        \end{cases}
%\vspace{-2mm}
\end{equation}

\noindent where $\mathit{CR} \in [0.0,1.0]$ controls the fraction of parameters that are copied to the trial solution. Note, the relation $j=j_{rand}$ ensures that the trial vector is different from the original solution $\mathbf{x}_{i}^{(t)}$.

Mathematically, a differential selection can be expressed as follows:

\begin{equation}
%\vspace{-1mm}
\label{eq:de_sel}
 \mathbf{x}_{i}^{(t+1)}=\begin{cases}
          \mathbf{w}_{i}^{(t)} &\text{if } f(\mathbf{w}_{i}^{(t)}) \leq f(\mathbf{x}_{i}^{(t)}), \\
		  \mathbf{x}_{i}^{(t)} &\text{otherwise}\,.
        \end{cases}
%\vspace{-2mm}
\end{equation}

In a technical sense, crossover and mutation can be performed in several ways in differential evolution. Therefore, a specific notation was used to describe the varieties of these methods (also strategies) generally. For example, 'rand/1/bin' denotes that the base vector is randomly selected, 1 vector difference is added to it, and the number of modified parameters in the mutant vector follows a binomial distribution. The other standard DE strategies (also ensemble strategies \textit{ES} in DE) are illustrated in Table~\ref{tab:strat}. 

\begin{table*}[!htb]
\begin{center}
\caption{Ensemble of DE-strategies}
%\small
\begin{tabular}{ | r | l | l | }
\hline
%Alg. & D & Value & Griewangk & Rosenbrock & Sphere & Rastrigin & Ackley \\ \hline
Nr. & Strategy & Expression \\ \hline
\hline
1 & Best/1/Exp & $x_{i,j}^{(t+1)}=best_{j}^{(t)}+F \cdot (x_{r1,j}^{(t)}-x_{r2,j}^{(t)})$ \\
2 & Rand/1/Exp & $x_{i,j}^{(t+1)}=x_{r1,j}^{(t)}+F \cdot (x_{r2,j}^{(t)}-x_{r3,j}^{(t)})$ \\
3 & RandToBest/1/Exp & $x_{i,j}^{(t+1)}=x_{i,j}^{(t)}+F \cdot (best_{i}^{(t)}-x_{i,j}^{(t)})+F \cdot (x_{r1,j}^{(t)}-x_{r2,j}^{(t)})$ \\
4 & Best/2/Exp & $x_{i,j}^{(t+1)}=best_{i}^{(t)}+F \cdot (x_{r1,i}^{(t)}+x_{r2,i}^{(t)}-x_{r3,i}^{(t)}-x_{r4,i}^{(t)})$ \\
5 & Rand/2/Exp & $x_{i,j}^{(t+1)}=x_{r1,i}^{(t)}+F \cdot (x_{r2,i}^{(t)}+x_{r3,i}^{(t)}-x_{r4,i}^{(t)}-x_{r5,i}^{(t)})$ \\
6 & Best/1/Bin & $x_{j,i}^{(t+1)}=best_i^{(t)}+F \cdot (x_{r1,i}^{(t)}-x_{r2,i}^{(t)})$ \\
7 & Rand/1/Bin & $x_{j,i}^{(t+1)}=x_{r1,j}^{(t)}+F \cdot (x_{r2,j}^{(t)}-x_{r3,j}^{(t)})$ \\
8 & RandToBest/1/Bin & $x_{j,i}^{(t+1)}=x_{i,j}^{(t)}+F \cdot (best_{i}^{(t)}-x_{i,j}^{(t)})+F \cdot (x_{r1,j}^{(t)}-x_{r2,j}^{(t)})$ \\
9 & Best/2/Bin & $x_{j,i}^{(t+1)}=best_i^{(t)}+F \cdot (x_{r1,i}^{(t)}+x_{r2,i}^{(t)}-x_{r3,i}^{(t)}-x_{r4,i}^{(t)})$ \\
10 & Rand/2/Bin & $x_{j,i}^{(t+1)}=x_{r1,i}^{(t)}+F \cdot (x_{r2,i}^{(t)}+x_{r3,i}^{(t)}-x_{r4,i}^{(t)}-x_{r5,i}^{(t)})$ \\
\hline
\end{tabular}
\normalsize
%\vspace{-5mm}
\label{tab:strat}
\end{center}
\end{table*}

\subsection{Random forest}
Leo Breiman introduced the random forests method in 2001~\cite{breiman2001random}. Random forests (RF) is an ensemble learning method, which can be used for classification as well as regression. Many decision trees are constructed during their training time. RF is used in many real-world application and is suitable for every classification. 

RF was also introduced into the optimization in~\cite{Fister:GECCO, Fister:Msc}, where the authors applied it in combination with a hybrid bat algorithm~\cite{fister2013hybrid}.

%\begin{figure*}[htb]  %[!htb]  %
%    \begin{center}
%        \includegraphics [scale=1.0]{forest.pdf}  %
%        \caption{Random forests}
%        \label{pic:Forest}
%    \end{center}
%%\vspace{-5mm}
%\end{figure*}

\subsection{Extremely randomized trees}
Extremely randomized trees (EXT) were introduced by Pierre Geurts, Damien Ernst and Louis Wehenkel in~\cite{geurts2006extremely}. They are tree-based ensemble method suitable for supervised classification and regression problems. The algorithm of growing EXT is similar to RF, but there are two differences:
\begin{itemize}
\item EXT do not apply the bagging procedure to construct a set of the training samples for each tree. The same input training set is used to train all trees.
\item EXT pick a node split very extremely (both a variable index and variable splitting value are chosen randomly), whereas RF finds the best split (an optimal one by variable index and variable splitting value) amongst a random subset of variables.
\end{itemize}

\subsection{Gradient boosting}
%tudi cite na http
Gradient boosting (GB) is a machine learning technique for regression problems, which produces a prediction model in the form of an ensemble of weak prediction models, typically decision trees~\cite{friedman2002stochastic,friedman2001greedy}. It builds the model in a stage-wise fashion like other boosting methods do, and it generalizes them by allowing optimization of an arbitrary differentiable loss function. The GB method can also be used for classification problems by reducing them to regression with a suitable loss function. Its advantages are:
\begin{itemize}
\item natural handling of data of mixed types,
\item predictive power,
\item robustness to outliers.
\end{itemize}

\subsection{Decision trees}
Decision Trees~\cite{breiman1993classification} (DT) is a non-parametric supervised learning method used for classification and regression. The goal is to create a model that predicts the value of a target variable by learning simple decision rules inferred from the data features~\cite{scikit-learn}.
DT has many advantages:
\begin{itemize}
\item it is simple to understand,
\item trees can be visualized,
\item the cost is logarithmic,
\item it is able to handle numerical and categorical data,
\item it is able to handle multi-output problems.
\end{itemize}

\subsection{Generalized linear model}
The Generalized linear model (LM) is a very interesting method that consists of several variants. In our experiments, ridge regression was used, also known as Tikhonov regularization~\cite{Tikhonov:1998}. This model solves a regression model, where the loss function is the linear least squared function, and regularization is given by the $l^2$-norm.

\section{The proposed DE algorithm}
The proposed DE algorithm (Algorithm~\ref{DE}) acts as follows: The DE population $P = \{x_{ij}\}$ for $i = 1 \ldots \mathit{NP} \wedge j = 1 \ldots \mathit{D}$, where $\mathit{NP}$ denotes the population size and $D$ the dimension of a problem, is initialized randomly. In each generation, a test set $T_i$ is created for each vector $\mathbf{x}_{i}$ using the ensemble strategies $\mathit{ES}$ (Table~\ref{tab:strat}). Then, a validation set $V_i$ consisted of vector $\mathbf{v_{i}}$ is created using the strategy 'rand/1/bin' from vector $\mathbf{x}_{i}$. Finally, a particular regression method is launched, that on the basis of both the defined set $T_i$ and $V_i$, and predefined input parameter set of the regression method $I_i$, builds the new regression vector $\mathbf{r}_i$. This regression vector enters into a struggle with the corresponding candidate solution $\mathbf{x}_i$ for a position within a new DE population. The optimization is terminated, when the number of generations reached $T_{\mathrm{max}}$.

\begin{algorithm}[H]
\caption{The proposed DE algorithm}
\label{DE}
\small
\begin{algorithmic}[1]
\STATE Initialize the DE population $\mathbf{x}_i = (x_{i1}, ... , x_{iD})$ for $i = 1 \ldots \mathit{NP}$ 
\STATE \textbf{repeat} 
\STATE \ \ \textbf{for} {$i=1$} \textbf{to} {$i \leq \mathit{NP}$}
\STATE \ \ \ \ Create test set $T$ on vector $\mathbf{x}_i$ using ensemble strategies $\mathit{ES}$;
\STATE \ \ \ \ Create validation set $\mathbf{v}_i$ by applying strategy 'rand/1/bin' on vector $\mathbf{x}_i$;
\STATE \ \ \ \ Build regression vector $\mathbf{r}_i$ by applying the regression method using $T$ and $\mathbf{v}_i$;
\STATE \ \ \ \ \textbf{if} ($f(\mathbf{r}_i) < f(\mathbf{x}_i)$) Insert $\mathbf{r}_i$ into $Q$;
\STATE \ \ \ \ \textbf{else} Insert $\mathbf{x}_i$ into $Q$;
\STATE \ \ \ \ \textbf{end if}  
\STATE \ \ \textbf{endfor}
\STATE \ \ $P = Q;$  
\STATE \textbf{until} (Termination condition meet)
\STATE Postprocess results and visualization
\end{algorithmic}
\normalsize
%\vspace{-2mm}
\end{algorithm}

The create test set function (Algorithm~\ref{Regression}) starts with an empty set $T=\{\emptyset\}$. For each strategy $s \in \mathit{ES}$, a vector $\mathbf{t}$ is created that is added to test set $T$.

\begin{algorithm}[H]
\caption{Create test set function}
\label{Regression}
\small
\begin{algorithmic}[1]
\STATE $T = \emptyset$;
\STATE \textbf{forall} {$s \in ES$} 
\STATE \ \ Create vector $\mathbf{t}$ using ensemble strategy $s$ on vector $\mathbf{x}_i$;
\STATE \ \ Add vector $\mathbf{t}$ to test set $T$;
\STATE \textbf{endfor}
\end{algorithmic}
\normalsize
%\vspace{-2mm}
\end{algorithm}

Note that the regression vector $\mathbf{r}_i$ is evaluated only once per generation, i.e., just before entering into the selection process (line 7 in Algorithm~\ref{DE}). On the other hand, calculating the regression vector depends on the statistical rules of the particular regression method. Interestingly, the proposed algorithm was wholly implemented in Python because the scikit-learn python library~\cite{scikit-learn} was used for implementing the regression methods. 

\section{Experiments and results}
The goal of the experimental work was to show how various regression methods influence the performances of ensemble strategies in DE algorithm. In line with this, the original DE algorithm was hybridized with regression methods, like RF, EXT, GB, DT, and GLM and tested on well-known functions of dimension $D=10$ taken from literature~\cite{yang2010test}. The task of function optimization is to find the minimum value of fitness function. All the selected functions had global optima at value zero. This function test suite can be seen in Table~\ref{tab:detail}.

\begin{table*}[htb]
\begin{center}
\caption{Test suite}
\small
%\footnotesize
\begin{tabular}{ | l | l | l | l |  }
\hline
$f$ & Function & Definition & Range  \\ \hline
\hline
$f_1$ & Rosenbrock & $\mathop{F(\mathbf{x})}=\sum_{i=1}^{D-1} 100\, (x_{i+1} - x_i^2)^2 + (x_i-1)^2$ & $-15.00\leq x_i \leq15.00$ \\
\hline
$f_2$ & Rastrigin & $\mathop{F(\mathbf{x})}=n*10 + \sum_{i=1}^{D}(x_i^2-10\cos(2\pi x_i))$ & $-15.00\leq x_i \leq15.00$ \\
\hline
$f_3$ & Sphere & $\mathop{F(\mathbf{x})}=\sum_{i=1}^{D}x_{i}^2$ & $-100.00\leq x_i \leq100.00$ \\
\hline
$f_4$ & Griewangk & $\mathop{F(\mathbf{x})}=-\prod_{i=1}^{D}\cos\left(\frac{x_i}{\sqrt{i}}\right)+\sum_{i=1}^{D}\frac{x_i^2}{4000}+1$ & $-600\leq x_i \leq600$ \\ \hline
$f_5$ & Ackley & $\mathop{F(\mathbf{x})}=\sum_{i=1}^{D-1}\left(20+e^{-20}e^{-0.2\sqrt{0.5(x_{i+1}^2+x_{i}^2)}}-e^{0.5(\cos(2\pi x_{i+1})+\cos(2\pi x_i))}\right)$ & $-32.00\leq x_i \leq 32.00$ \\
\hline
\end{tabular}
\normalsize
\vspace{-5mm}
\label{tab:detail}
\end{center}
\end{table*}

During our experimental work, the original algorithm was compared with the regression methods on ensemble strategies in DE. As a result, six independent runs were conducted. The control parameters of DE were set during the experiments as: $F=0.5$, $\mathit{CR}=0.9$, and $\mathit{NP}=10$. Optimization of the DE algorithm was terminated when $T_{max} = 1{,}000$. In other words, the number of fitness function evaluations was limited to $10{,}000$. Each run was performed 25 times. The RF and EXT used 40 estimators during the tests. 

The results of the experiments are presented in Table~\ref{tab:results}, where the best mean values are in bold. As can be seen from the table, the RF on ensemble strategies in the DE algorithm achieved the best results by optimizing the functions $f_2$--$f_5$, whilst optimizing the function $f_1$ was performed better using the EXT regression method. The worst results were reported by the original DE algorithm.

%random forests and extra trees - num of estimatos = 40
\begin{table*}[htb]
\begin{center}
\caption{The results of the experiments}
\small
\begin{tabular}{ | c | c | r | r | r | r | r | r | }
\hline 
%Alg. & D & Value & Griewangk & Rosenbrock & Sphere & Rastrigin & Ackley \\ \hline
Alg. & D & Value & $f_1$ & $f_2$ & $f_3$ & $f_4$ & $f_5$ \\ \hline
\hline
\multirow{5}{*}{DE}  &  \multirow{5}{*}{10}  &  Best  & 3.30E+003 & 6.70E+001 & 1.70E+002 & 5.30E-004 & 2.80E+000 \\
  &    &  Worst  & 4.80E+006 & 3.90E+002 & 4.00E+003 & 3.80E-001 & 1.60E+001 \\
  &    &  Mean  & 1.00E+006 & 1.90E+002 & 1.50E+003 & 5.60E-002 & 1.10E+001 \\
  &    &  Median  & 4.20E+005 & 1.70E+002 & 1.30E+003 & 2.70E-002 & 1.10E+001 \\
  &    &  StDev  & 1.30E+006 & 9.30E+001 & 1.00E+003 & 8.30E-002 & 2.70E+000 \\
\hline 
\multirow{5}{*}{DE+RF}  &  \multirow{5}{*}{10}  &  Best  & 8.79E+000 & 0.00E+000 & 0.00E+000 & 0.00E+000 & 4.44E-016 \\
  &    &  Worst  & 8.89E+000 & 0.00E+000 & 0.00E+000 & 0.00E+000 & 4.44E-016 \\
  &    &  Mean  & 8.87E+000 & \textbf{0.00E+000} & \textbf{0.00E+000} & \textbf{0.00E+000} & \textbf{4.44E-016} \\
  &    &  Median  & 8.87E+000 & 0.00E+000 & 0.00E+000 & 0.00E+000 & 4.44E-016 \\
  &    &  StDev  & 2.00E-002 & 0.00E+000 & 0.00E+000 & 0.00E+000 & 0.00E+000 \\
\hline 
\multirow{5}{*}{DE+EXT}  &  \multirow{5}{*}{10}  &  Best  & 7.77E-008 & 1.09E+001 & 4.44E-042 & 2.90E-001 & 3.99E-015 \\
  &    &  Worst  & 4.16E+000 & 3.19E+001 & 1.32E-038 & 2.90E-001 & 3.99E-015 \\
  &    &  Mean  & \textbf{2.77E+000} & 2.33E+001 & 2.57E-039 & 2.00E-001 & 3.99E-015 \\
  &    &  Median  & 2.72E+000 & 2.45E+001 & 1.50E-039 & 2.10E-001 & 3.99E-015 \\
  &    &  StDev  & 9.90E-001 & 4.47E+000 & 2.88E-039 & 4.90E-002 & 0.00E+000 \\
\hline 
\multirow{5}{*}{DE+GB}  &  \multirow{5}{*}{10}  &  Best  & 4.54E+000 & 1.43E+001 & 0.00E+000 & 9.00E-002 & 4.44E-016 \\
  &    &  Worst  & 7.73E+000 & 3.23E+001 & 0.00E+000 & 3.60E-001 & 4.44E-016 \\
  &    &  Mean  & 6.43E+000 & 2.12E+001 & 0.00E+000 & 2.30E-001 & 4.44E-016 \\
  &    &  Median  & 6.61E+000 & 2.07E+001 & 0.00E+000 & 2.20E-001 & 4.44E-016 \\
  &    &  StDev  & 8.00E-001 & 4.35E+000 & 0.00E+000 & 6.00E-002 & 0.00E+000 \\
\hline  
\multirow{5}{*}{DE+GB}  &  \multirow{5}{*}{10}  &  Best  & 1.29E+000 & 1.54E+001 & 0.00E+000 & 7.00E-002 & 4.44E-016 \\
  &    &  Worst  & 4.33E+000 & 2.80E+001 & 0.00E+000 & 3.40E-001 & 4.44E-016 \\
  &    &  Mean  & 2.81E+000 & 2.30E+001 & \textbf{0.00E+000} & 2.20E-001 & \textbf{4.44E-016} \\
  &    &  Median  & 2.90E+000 & 2.41E+001 & 0.00E+000 & 2.20E-001 & 4.44E-016 \\
  &    &  StDev  & 8.50E-001 & 3.62E+000 & 0.00E+000 & 6.00E-002 & 0.00E+000 \\
\hline  
\multirow{5}{*}{DE+DT}  &  \multirow{5}{*}{10}  &  Best  & 0.00E+000 & 0.00E+000 & 0.00E+000 & 1.10E-001 & 4.44E-016 \\
  &    &  Worst  & 8.90E+000 & 0.00E+000 & 0.00E+000 & 2.90E-001 & 4.44E-016 \\
  &    &  Mean  & 7.83E+000 & \textbf{0.00E+000} & \textbf{0.00E+000} & 2.10E-001 & \textbf{4.44E-016} \\
  &    &  Median  & 8.90E+000 & 0.00E+000 & 0.00E+000 & 2.10E-001 & 4.44E-016 \\
  &    &  StDev  & 2.89E+000 & 0.00E+000 & 0.00E+000 & 4.00E-002 & 0.00E+000 \\
\hline 
\multirow{5}{*}{DE+LM}  &  \multirow{5}{*}{10}  &  Best  & 4.54E+000 & 1.43E+001 & 0.00E+000 & 9.00E-002 & 4.44E-016 \\
  &    &  Worst  & 7.73E+000 & 3.23E+001 & 0.00E+000 & 3.60E-001 & 4.44E-016 \\
  &    &  Mean  & 6.43E+000 & 2.12E+001 & \textbf{0.00E+000} & 2.30E-001 & \textbf{4.44E-016} \\
  &    &  Median  & 6.61E+000 & 2.07E+001 & 0.00E+000 & 2.20E-001 & 4.44E-016 \\
  &    &  StDev  & 8.00E-001 & 4.35E+000 & 0.00E+000 & 6.00E-002 & 0.00E+000 \\
\hline
\end{tabular}
\normalsize
\vspace{-5mm}
\label{tab:results}
\end{center}
\end{table*}

\begin{SCfigure}[][htb]
  \centering
  \includegraphics[width=0.5\textwidth]%
    {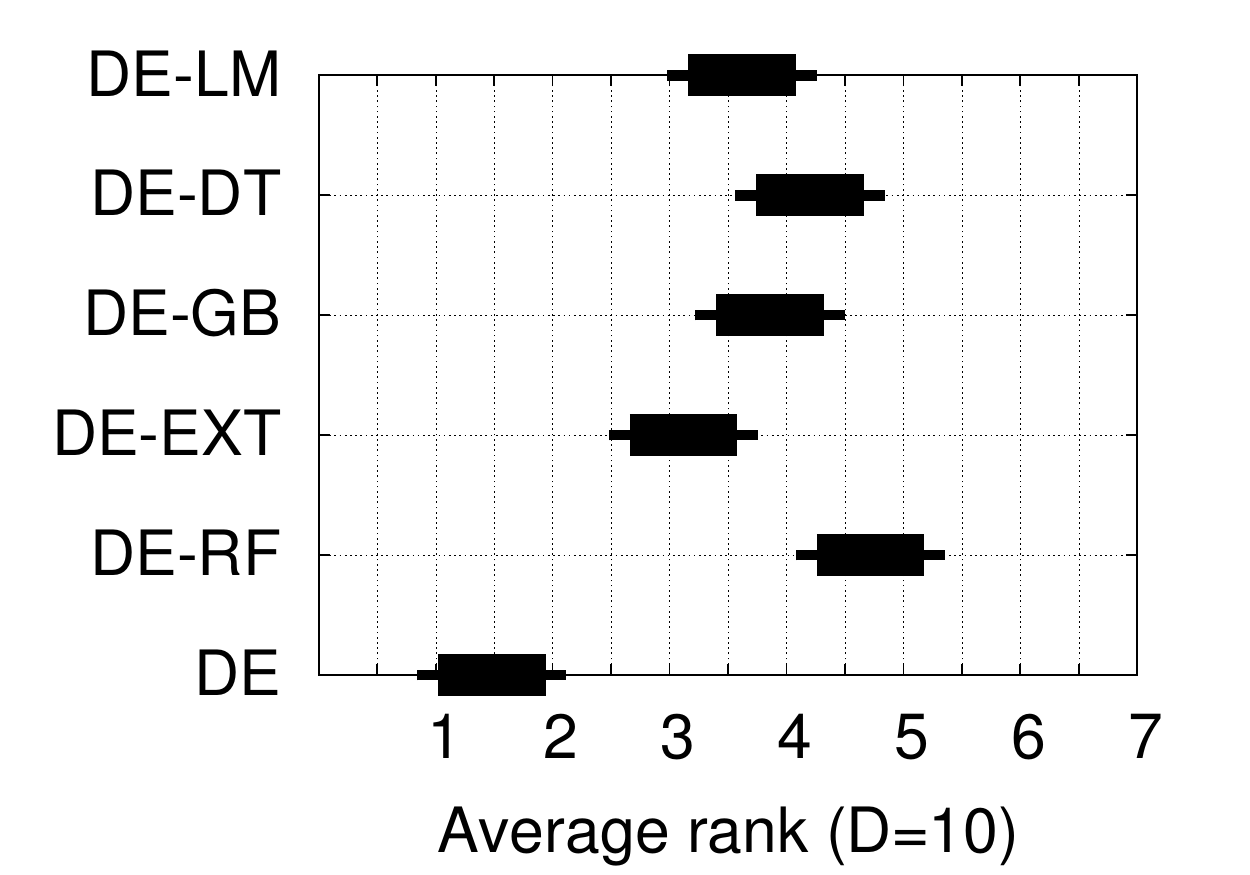}% picture filename
  \caption{Diagram shows the ranks and confidence intervals (critical differences) for the algorithms under consideration. Two algorithms are significantly different if their intervals do not overlap. The diagram represents the results of test functions with dimension $D=10$.}
  \label{fig:Sub_2}
\end{SCfigure}

A Friedman non-parametric test was conducted in order to show how good the obtained results were. The Friedman test~\cite{Friedman:1937} compares the average ranks of the algorithms. A null-hypothesis states that two algorithms are equivalent and, therefore, their ranks should be equal. If the null-hypothesis is rejected, i.e., the performance of the algorithms is statistically different, the Bonferroni-Dunn test~\cite{Demsar:2006} is then performed that calculates the critical difference between the average ranks of those two algorithms. When the statistical difference is higher than the critical difference, the algorithms are significantly different. The equation for the calculation of critical difference can be found in~\cite{Demsar:2006}. The results of the Friedman non-parametric test using the significance level $0.05$ are presented in Fig.~\ref{fig:Sub_2}.

As can be seen from Fig.~\ref{fig:Sub_2}, using any of the proposed regression method improved the results of the original DE significantly. Furthermore, using the RF regression method also improved the results of the EXT and LM regression methods. Interestingly, the results of the older non-ensemble learning method DT substantially outperformed the results of other regression methods, except RF.
 
\section{Conclusion}
This paper suggests using various regression methods on ensemble strategies in the DE algorithm. The best DE strategy was selected on the basis of statistical regression during the run. Each of the five used regression methods (i.e., RF, EXT, DT, LM, and GB) outperformed the results of the original DE algorithm. The best results were achieved by the RF regression method, which also significantly outperformed the results when using the EXT and LM regression methods. In summary, the promising results using the various regression methods on ensemble strategies in DE showed that a bright future could await this direction of DE development.

\bigskip{\small \smallskip\noindent Updated 2 July 2013.}
\end{document}